\documentclass{article}

\usepackage[accepted]{icml2024}

\usepackage[utf8]{inputenc} % allow utf-8 input
\usepackage[T1]{fontenc}    % use 8-bit T1 fonts
\usepackage{hyperref}       % hyperlinks
\usepackage{url}            % simple URL typesetting
\usepackage{booktabs}       % professional-quality tables
\usepackage{amsfonts}       % blackboard math symbols
\usepackage{nicefrac}       % compact symbols for 1/2, etc.
\usepackage{microtype}      % microtypography
\usepackage{xcolor}         % colors

\usepackage{algorithm}
\usepackage{algorithmic}
\usepackage{amsmath,amssymb}
\usepackage{bm}
\usepackage{graphicx}
\usepackage{caption}
\usepackage{subcaption}
\usepackage{multirow}

\usepackage{comment}
\usepackage{tablefootnote}

\usepackage{libraries/mkolar_definitions}

\usepackage[turnon]{libraries/mynotes}

\newcommand{\policy}{\pi_\theta}
\newcommand{\expert}{\pi_e}

\newcommand{\traje}{\tau}

\newcommand{\demo}{\Dcal_\text{E}}

\newcommand{\som}{\rho_\pi(s)}
\newcommand{\somf}{\sum_{t=0}^{\infty} \gamma^t P(s_t = s | \pi)}

\newcommand{\saom}{\rho_\pi(s,a)}
\newcommand{\saomf}{\rho_\pi(s)\pi(a|s)}
\newcommand{\saeom}{\rho_{\pi_{\text{E}}}(s,a)}

\newcommand{\ssom}{\rho_\pi(s,s')}
\newcommand{\ssomf}{\int_\Acal \rho_\pi(s,\bar{a})\Tcal(s'|s,\bar{a}){\rm d}\bar{a}}
\newcommand{\sseom}{\rho_{\pi_{\text{E}}}(s,s')}

\newcommand{\invom}{\rho_\pi(a|s,s')}
\newcommand{\inveom}{\rho_{\pi_{\text{E}}}(a|s,s')}

\newcommand{\lbc}{\Lcal_{\mathrm{BC}}}

\newcommand{\divergence}{\mathrm{D}}
\newcommand{\kld}{\mathrm{D}_{\mathrm{KL}}} 
\newcommand{\jsd}{\mathrm{D}_{\mathrm{JS}}}

\definecolor{ForestGreen}{RGB}{76, 175, 80}

\newcommand{\mytitle}{Mimicking Better by Matching the Approximate Action Distribution}
\newcommand{\mytitleshort}{MAAD}
\newcommand{\mytitlerunning}{Mimicking Better by Matching the Approximate Action Distribution}

\newcommand{\ilo}{ILO}
\newcommand{\ild}{ILD}

\newcommand{\refeq}[1]{Eq.~\ref{#1}}
\newcommand{\refsec}[1]{Section~\ref{#1}}
\newcommand{\reffig}[1]{Fig.~\ref{#1}}
\newcommand{\reftab}[1]{Table~\ref{#1}}

\newcommand{\refalg}[1]{Algorithm~\ref{#1}}

\icmltitlerunning{\mytitlerunning}

\begin{document}

\twocolumn[
\icmltitle{\mytitle}




\begin{icmlauthorlist}
\icmlauthor{João~A. Cândido~Ramos}{unige,hesso}
\icmlauthor{Lionel Blondé}{hesso}
\icmlauthor{Naoya Takeishi}{unitokyo,riken}
\icmlauthor{Alexandros Kalousis}{hesso}
\end{icmlauthorlist}

\icmlaffiliation{unige}{University of Geneva (UNIGE), Switzerland}
\icmlaffiliation{hesso}{University of Applied Sciences and Arts Western (HES-SO), Switzerland}
\icmlaffiliation{unitokyo}{The University of Tokyo, Japan}
\icmlaffiliation{riken}{RIKEN Center for Advanced Intelligence Project, Japan}

\icmlcorrespondingauthor{João~A. Cândido~Ramos}{joao.candido@etu.unige.ch}

\icmlkeywords{Machine Learning, ICML}

\vskip 0.3in
]



\printAffiliationsAndNotice{} % otherwise use the standard text.

\begin{abstract}

In this paper, we introduce \mytitleshort, a novel, sample-efficient on-policy algorithm for Imitation Learning from Observations. \mytitleshort~utilizes a  surrogate reward signal, which can be derived from various sources such as adversarial games, trajectory matching objectives, or optimal transport criteria.
To compensate for the non-availability of expert actions, we rely on an inverse dynamics model that infers plausible actions distribution given the expert's state-state transitions; we regularize the imitator's policy by aligning it to the inferred action distribution. 
\mytitleshort~leads to significantly improved sample efficiency and stability. We demonstrate its effectiveness in a number of MuJoCo environments, both int the OpenAI Gym and the DeepMind Control Suite. We show that it requires considerable fewer interactions to achieve expert performance, outperforming current state-of-the-art on-policy methods. Remarkably, \mytitleshort~often stands out as the sole method capable of attaining expert performance levels, underscoring its simplicity and efficacy.

\end{abstract}

\section{Introduction}

\begin{figure}[ht]
    \centering
    \begin{subfigure}{\linewidth}
        \includegraphics[width=\linewidth]{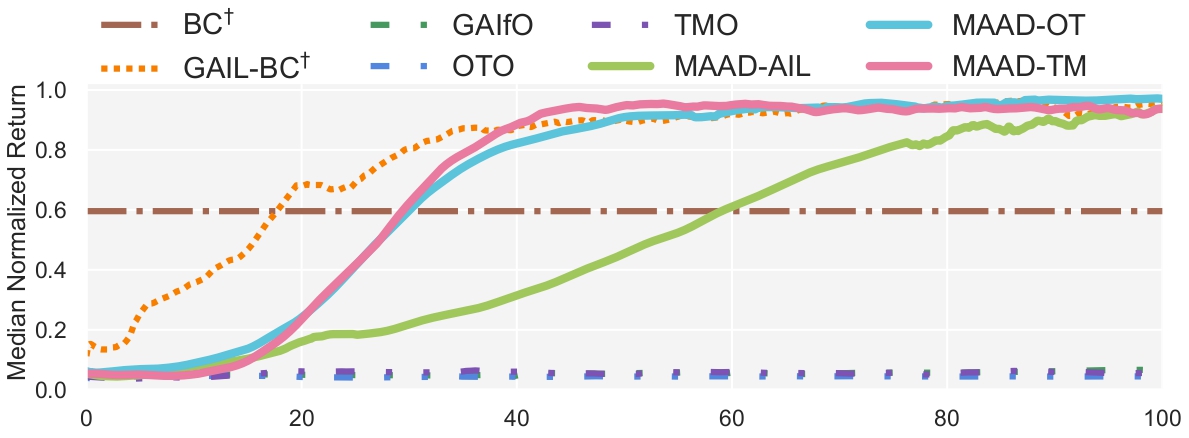}
        \vskip -0.05in
        \caption{DMC Suite}
        \label{fig:median_dmc}
    \end{subfigure}
    \begin{subfigure}{\linewidth}
        \includegraphics[width=\linewidth]{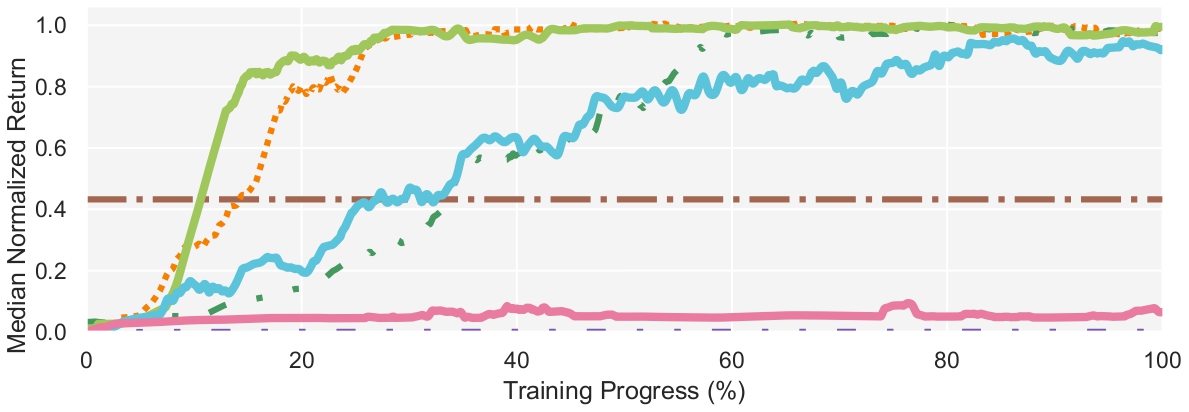}
        \vskip -0.05in
        \caption{OpenAI Gym}
        \label{fig:median_gym}
    \end{subfigure}
    \vskip -0.1in
    \caption{\textbf{Median Normalized Return}, over different environments, of various instantiations of our method (solid lines) versus baselines (dashed curves).
    %This plot is derived by quantalising training curves using a fixed number of quantiles, with median computation per algorithm across the different environments of each suite tested.
    Methods marked with $\dag$ have access to expert actions representing the best possible performance, all others \emph{do not}. More details on the construction of the figure in \refsec{sec:median_norm}.
    }
    \label{fig:median_dmc}
\end{figure}

Reinforcement learning (RL) trains agents to perform tasks by learning from rewards.%, a core approach in AI that fosters novelty and creativity in artificial behavior. 
Crafting a reward function that accurately reflects the intended task remains challenging \citep{Silver2017MasteringTG,Ouyang2022TrainingLM}. An alternative strategy involves teaching agents through demonstration, known as Imitation Learning (IL), where the goal is for the agent to closely replicate expert demonstrations, usually given as sequences of state-action pairs. While IL circumvents the complexity of designing reward functions, it requires access to expert actions, precluding its use in scenarios where actions are not observable, such as in motion capture data or video recordings.

Imitation learning from observations (\ilo), as opposed to imitation learning from demonstrations (\ild), focuses on learning from states alone, allowing agents to imitate expert behavior without accessing their actions. 
This approach is more versatile, addressing the limitations of IL in many practical contexts, and has gained significant interest for its broader applicability in fields like autonomous driving.
To facilitate imitation learning without direct action information, one strategy modifies algorithms to operate over state-state transitions rather than state-action pairs. This adaptation, as explored by \cite{Torabi2018GenerativeAI}, leverages state-state transitions to distinguish between expert and agent. A different strand of research focuses on inferring missing actions through inverse models of the environment, facilitating the application of traditional IL methods \citep{Hanna2017StochasticGA,Nair2017CombiningSL,Pavse2019RIDMRI,al2023ls}. 

Recent studies in \ild~\citep{Fujimoto2021AMA,Jena2020AugmentingGW,efficientimitate,Haldar2022WatchAM} have shown that integrating online IL techniques with offline Behavioral Cloning (BC) \citep{Pomerleau91,Bain1995AFF} improves sample efficiency by leveraging the strengths of both. Online IL methods learn through environmental interactions, whereas BC operates offline, making the interaction cost dependent solely on the online IL component.
Another angle to improve sample efficiency is to adopt an online yet off-policy training strategy \citep{Blond2018SampleEfficientIL,KostrikovNT20}.
In this context, \cite{Zhu2021OffPolicyIL} propose combining an off-policy algorithm with a BC regularizer, employing an inverse model to address the absence of actions.
However, off-policy approaches bring challenges such as extrapolation error and distributional shift \citep{Fujimoto2018OffPolicyDR,Fu2019DiagnosingBI,Levine2020OfflineRL}.
Instead, we propose an on-policy method for sample-efficient training in \ilo~that is both easy and stable to train.

Our approach combines a model-free base that dynamically infers rewards from environmental interactions, allowing for the use of different reward inference methods such as adversarial imitation, trajectory matching, and optimal transport, with a model-based regularizer that incorporates an inverse model of the environment.
The inverse modeling task estimates the posterior distribution of actions that are plausible within the simulator's physics, based on state transitions. The regularizer constrains the policy to align its actions closely with the action posterior distribution given by the inverse model as this is applied on expert demonstrations. %This forms a secondary objective of action matching, where the policy's actions are constrained to match those approximated from the inverse model applied on expert demonstrations, similar to a BC objective.
This forms a secondary objective of action matching, which is similar to a BC objective.

We refer to our method as \mytitle~(\mytitleshort).
By guiding the policy towards selecting actions that comply with the physics intricately encoded within the simulator, we provide the agent with richer supervisory information throughout the training process.
This results in significant sample efficiency gains; our policies train much faster than all competitive baselines, and they even train in settings where most of the other baselines fail.
We demonstrate the superiority of our method on a spectrum of complex continuous control tasks developed with the MuJoCo physics engine \citep{todorov2012mujoco}.
We tackle the ones distributed in OpenAI Gym \citep{brockman2016openai} and in the DeepMind Control Suite \citep{tunyasuvunakool2020}.

\section{Related Work}
Interest in Imitation Learning from Observations has grown in recent years, as it enables imitation from a variety of sources where actions are not explicit. Recent advancements have expanded the range of possibilities for reward approximation, such as adversarial methods, trajectory matching techniques, and optimal transport theory. In this work, we explore how to combine these with a behavior cloning-like regularizer that typically requires access to actions.

One strand of works for reward approximation adopts adversarial-based approaches
%, which build on GANs \citep{Goodfellow2014GenerativeAN}, 
and learn a discriminative function (discriminating between expert and agent data) that serves as a surrogate for the reward function. 
%IL methods following this approach are based on GAIL \citep{Ho2016GenerativeAI}. 
GAIL, \citep{Ho2016GenerativeAI}, is probably the most prominent example of adversarial-based methods for the \ild~setting, relying on state-action pairs to discriminate between the expert and the agent.
GAIfO, \cite{Torabi2018GenerativeAI}, adapts GAIL to the \ilo~context by using state transitions instead of state-action pairs. 
%. It does this by substituting the state-action pairs in GAIL, which differentiate between agent and expert data, with state-state transitions. 
\citet{Yang2019ImitationLF} have shown that GAIL and GAIfO are connected by the inverse dynamics disagreement; a divergence measure between the inverse dynamics models of the expert and the agent.
% VAIfO \citep{Peng2018VariationalDB} tackles the problem of instability in adversarial training by introducing a variational discriminator bottleneck.
Additionally, there have been efforts to adapt the GAIL framework to an off-policy setting, as seen in works by \cite{Blond2018SampleEfficientIL} and \cite{Kostrikov2018DiscriminatorActorCriticAS}.

Another simple, but sometimes surprisingly effective, approach to reward approximation infers rewards based on trajectory matching distances; such works basically compute euclidean distances between the expert and agent sequences and use these as rewards \citep{Englert2013ModelbasedIL,Peng2018DeepMimicED}.  
Finally, another approach to inferring surrogate rewards  relies on Optimal Transport (OT). Optimal transport methods establish distribution alignments (joint distributions) that optimize some cost function when marginalized over the joint distribution. They are used in imitation learning to guide the agent's distribution towards the expert's distribution.
For instance, SIL \citep{Papagiannis2020ImitationLW} uses Sinkhorn \citep{NIPS2013_af21d0c9}, PWIL \citep{Dadashi2020PrimalWI} employs greedy Wasserstein, GDTW-IL is based on Gromov Dynamic Time Warping \citep{cohen2022imitation}, and GWIL \citep{Fickinger2021CrossDomainIL} incorporates Gromov-Wasserstein \citep{Peyr2016GromovWassersteinAO}.
Both trajectory matching and optimal transport approaches can also be easily deployed to the \ild~or the \ilo~setting by operating over state-action, state-state or state-only sequences.

Adversarial- and optimal transport-based methods for imitation learning from demonstrations have been coupled with (regularized by) behavioral cloning, resulting in significant sample efficiency improvements compared to non-regularized baselines. GAIL-BC, \citep{Jena2020AugmentingGW}, and ROT, \citep{Haldar2022WatchAM}, regularize GAIL and SIL with behavioral cloning, respectively. Additional IL approaches that improve sample efficiency through BC include \citep{Fujimoto2021AMA, efficientimitate}. However, extending these sample efficiency gains to the \ilo~setting poses challenges due to the absence of actions.

Within the \ilo~setting a number of works take an alternative approach to address the non-availability of actions; instead of operating over state-state transitions they seek to infer the non-available actions. They do so by learning an inverse dynamics model, use it to infer the actions and continue with the application of standard \ild~methods \citep{TorabiWS18, Hanna2017StochasticGA, Nair2017CombiningSL, Pavse2019RIDMRI, al2023ls, Radosavovic2020StateOnlyIL}.
Additionally, it can also be used to pretrain a policy \citep{NEURIPS2023_d36dfcdb}.
The availability of an inverse dynamics model that can account for the non-available actions opens the door for bringing the sample efficiency properties of behavioral cloning within the \ilo~context. \cite{Zhu2021OffPolicyIL} follow such a path combining an off-policy imitation learning algorithm \cite{Kostrikov2018DiscriminatorActorCriticAS} with a BC-like regularizer and demonstrate sample efficiency gains.
Similarly, Hybrid-RL \citep{Guo2019HybridRL} merges standard RL with a BC-like regularizer; however, it assumes access to the expert's reward function, which limits its broader applicability. 

Our work brings the sample efficiency gains of behavioral-cloning regularization to on-policy imitation learning from observations by using inverse dynamics models to regularize the agent's policy; we demonstrate its applicability with a diverse range of reward surrogate learning methods and show that it learns to imitate even in settings where competing methods simply fail.
%Our research broadens the scope of existing frameworks by integrating on-policy approaches, with a diverse array of reward surrogate formulations, and BC-like regularizer into the \ilo~setting.
\section{Background}\label{sec:background}

\paragraph{Markov Decision Process (MDP)}
We will consider agents in a $\gamma$-discounted infinite horizon Markov decision process (MDP) $\Mcal = \langle \Scal, \Acal, \Tcal, \rho_0, r, \gamma \rangle$, where $\Scal$ and $\Acal$ are the state and action space respectively, $\Tcal: \Scal \times \Acal \times \Scal \to \mathbb{R}_{\geq 0}$ is the transition distribution, $\rho_0: \Scal \to \mathbb{R}_{\geq 0}$ is the initial state distribution, $r: \Scal \times \Acal \rightarrow \mathbb{R}$ is the reward function and $\gamma$ is the discount factor.

\paragraph{Demonstrations and Observations} 
In the \ild~setting, we have an expert $\expert$ which generates trajectories. A trajectory, $\tau$, is a sequence of state-action pairs, $\tau = \{ (s_0, a_0), (s_1, a_1), \dots, (s_T, a_T) \}$,  collected during one episode. We have access to a set of demonstrations $\mathcal{D}_E = \{ \tau_i \}, \tau_i \thicksim \expert$ on which we train our imitation policy. In contrast, in the \ilo~setting, we do not have access to actions. Here, a trajectory $\zeta$ is a sequence of state-state transitions, $\zeta = \{ (s_0, s_1), (s_1, s_2), \dots, (s_{T-1}, s_T) \}$. As such, in this work, we train our imitation policy on the set of observations $\mathcal{D}_E = \{ \zeta_i \}, \zeta_i \thicksim \expert$.

\paragraph{Behavior Cloning} BC, \citep{Pomerleau91,Bain1995AFF}, tackles imitation learning using supervised learning. The policy is trained to maximize the likelihood of the expert's actions: $\lbc =  - \EE_{s,a \sim \demo} [ \log(\policy(a|s)) ]$. BC being basically a supervised learning method, it does not have an exploration mechanism. This makes the policy subject to compounding errors \citep{Ross2010ARO} and suboptimal asymptotic performance. BC also assumes the presence of expert state-action pairs in the demonstration set $\demo$. To address the \ilo~setting in which we do not have actions, \cite{TorabiWS18} learn an inverse dynamics model, $p(a|s,s')$, which they use to infer actions and they follow with a standard BC application. They train the inverse dynamics model on a dataset of $(s,a,s')$ triplets collected with a random policy. Such an inverse model provides access to actions that are physically plausible with the given $(s,s')$ transitions; these actions will typically be a superset of the actions that the expert would have chosen. 

\paragraph{Occupancy Measure}
%The occupancy measure (OM) \citep{Ho2016GenerativeAI} is a statistical measure that  describes the distribution of time spent in different states of a Markov decision process (MDP) under a given policy $\pi$. 
The occupancy measure (OM) can be thought of as the distribution of states that are encountered in a Markov Decision Process (MDP) under a given policy $\pi$ \citep{Ho2016GenerativeAI}. 
The state OM, the most basic type of OM, is the discounted sum of the stationary state probability density, calculated over time for a given policy $\som = \somf$.
We can use the state OM to define additional occupancy measures that are applicable to different supports within an MDP.
These measures include the state-action occupancy measure $\saom = \saomf$, the state-state transition occupancy measure $\ssom = \ssomf$, and the density function of the inverse dynamics model under the policy $\pi$, $\rho_\pi(a|s, s')$ which is defined as follows:
\begin{equation}\label{eq:inversemodel}
  \rho_\pi(a|s, s') := \frac{\mathcal{T}(s'|s, a)\pi(a|s)}{\int_\mathcal{A} \mathcal{T}(s'|s, \bar{a})\pi(\bar{a}|s)d\bar{a}}
\end{equation}
where $\mathcal{T}(s'|s, a)$ is the probability to transition to the next state $s'$, given by the environment.

\paragraph{Generative Adversarial Imitation Learning}  GAIL, \citep{Ho2016GenerativeAI}, minimizes the Jensen–Shannon divergence between the agent $\saom$ and the expert $\saeom$ occupancy measures. The learned policy is thus given by $\argmin_\pi \jsd( \saom || \saeom)$. In practice, this minimization is achieved by training a discriminator and a policy in an adversarial manner where the discriminator provides a proxy reward to an on-policy reinforcement learning algorithm, such as PPO \citep{Schulman2017ProximalPO}, that trains the policy. 
\cite{Torabi2018GenerativeAI} propose to replace the state-action occupancy measure with the state transition OM, thus the learned policy is given by $\argmin_\pi \jsd(\ssom || \sseom)$, and thus eliminates the need for actions.

\paragraph{Trajectory Matching and Optimal Transport for Imitation Learning}
Trajectory matching involves aligning the agent's trajectory ($\xi^{\pi_{\theta}}_i$) with the expert's trajectory ($\zeta^E_i$) to learn desired behaviors. This process typically employs a similarity measure to quantify how closely the agent's states follow those of the expert. One common approach is using a cost function, such as Euclidean or cosine distance, to evaluate the difference at each point between $\xi^{\pi_{\theta}}_i$ and $\zeta^E_i$. The goal is to minimize this cost, guiding the agent towards replicating the expert's trajectory as closely as possible. 
More recently, optimal transport (OT)-based techniques have been proposed in imitation learning \citep{Dadashi2020PrimalWI,Papagiannis2020ImitationLW,cohen2022imitation, Haldar2022WatchAM}.
These methods focus on assessing the proximity between expert trajectories ($\zeta^E_i$) and agent trajectories ($\xi^{\pi_{\theta}}_i$) by evaluating the optimal transfer of probability mass from $\xi^{\pi_{\theta}}_i$ to $\zeta^E_i$. The surrogate reward for an observation is calculated as $r(s_t) = - \sum_{t'=1}^{T} C_{t,t'} \mu^{*}_{t,t'}$, where $C_{t,t'}$ is a cost matrix $C_{t,t'} = c(s_t, s^e_{t'})$, determining the cost of aligning a state from the agent's trajectory $s$ with a state from the expert's trajectory $s^e$. The term $\mu^{*}$ represents the optimal alignment between these trajectories.

% \begin{equation}\label{eq:ot}
% r(s_t) = - \sum_{t'=1}^{T} C_{t,t'} \mu^{*}_{t,t'}
% \end{equation}
\section{Model}

We will now introduce \mytitleshort~for Matching Approximate Action Distributions.
As is typical in imitation learning, we aim to establish a policy $\policy$ that will closely mimic the expert's policy $\expert$. In imitation learning from observations, this can be achieved by minimizing the discrepancy between the expert's and agent's state occupancy measures. This can be viewed as an alignment of the probability distributions over states that the expert and agent visit during their respective interactions with the environment.
This corresponds to the following objective function:
\begin{equation}\label{eq:ilo}
  \Lcal_{\text{policy}} = \divergence (\ssom || \sseom)
\end{equation}
In the context of imitation learning from demonstrations where the actions carried out by the expert are available, approaches such as \citep{Fujimoto2021AMA, Jena2020AugmentingGW, efficientimitate, Haldar2022WatchAM} have shown that regularizing the imitator policy with a behavioral cloning objective significantly improves the convergence rate. Driven by this recurring result, we seek to also benefit from these advantages in the \ilo~setting by constraining our policy with such an auxiliary objective.
Assuming for a moment that expert actions were available, we can define the following behavioral cloning loss term:
\begin{equation}\label{loss/bc}
  \Lcal_{\text{BC}} = -\EE_{s,a\sim\traje, \traje \sim \expert}[\log(\policy(a | s))]
\end{equation}
which provides additional supervision to the learned policy by urging it to assign high probability to the actions that the expert selected in the demonstration dataset.
We can thus extend the objective function in \refeq{eq:ilo} with this loss term, resulting in the following regularized objective:
\begin{equation}\label{eq:loss_gailbc}
  \mathcal{L} = \Lcal_{\text{policy}} + \lambda\Lcal_{\text{BC}}
\end{equation}
The equation \refeq{eq:loss_gailbc} is a generalization of existing BC-regularized IL algorithms. For instance, by substituting the state-state occupancy measure in \refeq{eq:ilo} with a state-action occupancy measure, we obtain the GAIL-BC algorithm \citep{Jena2020AugmentingGW}. On the other hand, if $\Lcal_{\text{policy}}$ is based on TD3 \citep{Fujimoto2018AddressingFA}, the resulting approach is TD3-BC \citep{Fujimoto2021AMA}. Furthermore, if we introduce an OT-based trajectory matching approach for the state occupancy measure, we obtain the ROT algorithm\footnote{ROT has BC pretraining phase, which we omit here.} \citep{Haldar2022WatchAM}. 
As is obvious, this is applicable only to the \ild~setting since the expectation in \refeq{loss/bc} is taken with respect to actions (and states) sampled from the expert.

If we are to use a loss term similar to the one in \refeq{loss/bc} to provide additional supervision, then we need a way to define what would be meaningful actions to select given that we do not have such information available from the expert data. The approach we take here is to learn the posterior distribution of actions that are physically plausible, under the used simulator, given a state-state transition, i.e. we want to approximate the true posterior $p(a | s, s')$. What we have here is basically an instance of what is known as simulation-based inference \citep{Cranmer2020}, and in particular amortized posterior inference \citep{papamakarios2016fast, lueckmann2017flexible, greenberg2019automatic}.

Learning the posterior distribution means actually solving the inverse dynamics problem of determining which are the actions that could have produced the observed $(s,s')$ transition. We should stress here that these are not necessarily the actions that the expert would have chosen, they are rather a superset of the possible expert actions, since they are all plausible actions under the $(s,s')$ transition. However, even though they are potentially a superset of the expert actions, they nevertheless provide considerable supervisory information since they will guide the expert to select actions that are physically plausible. Such guidance is particularly valuable since the gradients that we get from the~\refeq{loss/bc} loss are much more informative than the ones that we get from the policy gradient optimization, since it is a supervised learning problem.

\subsection{Inverse Dynamics Model (IDM)}\label{sec:idm}

Very often in inverse problems there is not a single inverse solution but rather a set of solutions and the respective posterior distribution has a multi-modal structure \citep{Ardizzone2018AnalyzingIP}. We thus choose to give the more general formulation to model the posterior distribution as a mixture density network (MDN) \citep{Bishop1994MixtureDN} which by design models multi-modality; the learned posterior distribution is given by:
\begin{equation}\label{eq:inv_def}
  p_{\alpha, \psi}(a|s,s') = \sum_{k=0}^{K-1} \alpha_k(s, s') \psi_k(a|s, s')
\end{equation}
which is a mixture of $K$ distributions $\psi_k(a|s, s')$ where $\alpha_k(s,s')$ is the probability of picking the $k$-th component of the mixture for the state transition $(s,s')$.
We use Gaussians as the mixture components. The number of components $K$ is a hyperparameter, which depends on the environment. Despite having experimented with values of $K>1$ in preliminary sweeps, we found that for the considered suites of environments, $K=1$ was already yielding satisfactory results. This indicates that for both suites, leveraging the potential multimodality of inverse models was not instrumental, further strengthening the simplicity and robustness of the method.

To train the IDM model, we need to collect $(s,a,s')$ triplets; a naive approach would be to collect such triplets using one or more random policies, but this would explore only a very small part of the joint space. Instead, we interleave the IDM training with the policy training. We take advantage of the interactions with the environment that take place as a result of learning online and push the collected triplets to a replay buffer $\Rcal$. We train the inverse model $p_{\alpha, \psi}$ on the data from the replay buffer until convergence, update the policy and repeat the process. We warm-start the inverse model from the one obtained in the previous step. Note that the replay buffer will contain the $10^5$ most recent samples from different policies, starting from a random policy to the currently established policy. The inverse model is learned over all these data points. Even though these data points are obtained from different policies, the learned inverse model will be valid for all of them since it reflects the underlying physics implemented in the simulator, which are independent of the policy.
We use the negative log-likelihood as the training objective for the inverse model:
\begin{equation}\label{eq:inv_loss}
  \Lcal_{\text{inv}} = - \EE_{(s,a,s') \sim \Rcal}[\log(p_{\alpha, \psi}(a|s,s'))]
\end{equation}

\subsection{Controlling the Policy Learning with the Inverse Dynamics Model}

Our model has two basic components, a policy learning component which minimizes the loss given in \refeq{eq:ilo} and an inverse model learning component which will be learned using the loss in \refeq{eq:inv_loss}. We will use the inverse model to further guide the policy learning with the help of a behavioral cloning loss similar to \refeq{loss/bc}.

The policy learning component is basically an RL algorithm which minimizes \refeq{eq:ilo} by learning the policy $\policy$ using Proximal Policy Optimization (PPO) \citep{Schulman2017ProximalPO} where the surrogate rewards can come from different sources.
In this paper, we will explore three different approaches for generating surrogate rewards.
Firstly, we instantiate \mytitleshort-AIL. This method obtains rewards through an adversarial training mechanism, similar to GAIfO \cite{Torabi2018GenerativeAI}, where the agent's goal is to mimic the expert's behavior closely enough to fool a discriminator trained to distinguish between them. This implies training an additional model, namely the discriminator. We train the discriminator $D_\phi$ using the cross-entropy loss on $(s,s')$ transitions sampled from the policy and the expert. The reward proxy that we use to train the policy is $r(s, s') = - \log(1-D(s,s'))$.
Secondly, we explore the trajectory matching through \mytitleshort-TM. In this approach, the agent's and expert's trajectories are considered to be aligned and of the same length, and the reward is calculated based on the Euclidean distance between the agent's current state and the corresponding expert state at each step. This method is geared towards closely following the expert's trajectory step by step. This method is naive and does not account for misalignment in the trajectories. It is therefore the fastest to compute because it relies on a simple distance that does not involve an extra model nor require an extra algorithm to be run.
Lastly, we investigate an Optimal Transport (OT) based method, \mytitleshort-OT. Here, rewards are obtained using the Sinkhorn algorithm \citep{NIPS2013_af21d0c9}, similar to what was proposed in SIL \cite{Papagiannis2020ImitationLW} and ROT \cite{Haldar2022WatchAM}, with a cost matrix established on the cosine distance between states. This method is focused on minimizing the overall cost of transforming the agent's state distribution to that of the expert, thereby assessing the similarity of entire trajectories. Note, the Sinkhorn algorithm needs to be run for each trajectory, and is expensive to run.

\begin{table}[h]
\centering
\small % Reduce font size to fit the table in a single column
\begin{tabular}{p{0.38\linewidth}|p{0.18\linewidth}|p{0.28\linewidth}}
\mytitleshort-AIL & \mytitleshort-TM & \mytitleshort-OT \\
\hline
$r(s_t, s_{t+1}) =$ & $r(s_t) =$ & $r(s_t) =$ \\
$- \log(1-D(s_t,s_{t+1}))$ & $\| s_t - s_t^e \|$ &  $- \sum_{t'=1}^{T} C_{t,t'} \mu^{*}_{t,t'}$ \\
\end{tabular}
\caption{Comparison of \mytitleshort-AIL, \mytitleshort-TM, and \mytitleshort-OT reward formulations}
\label{tab:reward}
\end{table}

The second component of our loss quantifies the similarity between the actions produced by the policy over expert states and the IDM predictions, using expert state-state transitions. The inverse model is represented by the learned posterior distribution $p_{\alpha,\psi}(a|s,s')$. Given the policy and the inverse model, it becomes imperative to select a suitable behavioral cloning loss. An option to consider is a likelihood-based loss similar to that described in Equation \ref{loss/bc}. However, in this context, instead of sampling state-action pairs from the expert, we would sample  $(s,s')$ transitions. These transitions are then utilized to sample plausible actions from the learned inverse model, aiming to maximize the likelihood of these sampled actions under the policy.
\begin{equation*}
  - \EE_{(s,s') \sim \zeta, \zeta \sim \expert}
  \EE_{ a \sim p_{\alpha,\psi}(a|s, s')} \log \policy (a | s)
\end{equation*}
Another alternative is to minimize the (forward) KL divergence of the inverse model and the learned policy, i.e.
\begin{equation}\label{eq:reg}
  \Lcal_{\text{reg}} = \EE_{(s,s') \sim \zeta, \zeta \sim \expert} \kld(p_{\alpha,\psi}(a|s,s') || \pi_\theta(a|s))
\end{equation}
the two approaches are equivalent up to an entropy term of the inverse model.

The forward KL induces a mode covering behavior. In our setting, this divergence will punish the policy for not assigning weight to actions that have non-zero density under the inverse model. Conversely, the reverse KL will punish the policy from having a non-zero density over actions for which the inverse model places none. In that respect, the mode-seeking effect induced by the reverse KL divergence should be the desired behavior, since it will push the learned policy to converge to one of the modes of the inverse model, and not fall in between the several potential modes discovered by the learned multi-modal inverse model.
Considering our focus on sample efficiency, and given that preliminary experiments showed a $k=1$ to be sufficient, we chose to use the forward KL divergence to constrain the learned policy, similar to the approach in \cite{Zhu2021OffPolicyIL}.
The forward KL divergence converges faster to the target distribution, i.e., the inverse model, thereby presenting a considerable advantage. Conversely, the reverse KL divergence necessitates extensive exploration to establish a reasonable policy, which is challenging during early training stages.
Thus, the objective that we use to train the policy is:
\begin{equation}
\begin{split}
\Lcal = & \underbrace{\divergence (\ssom || \sseom)}_{\Lcal_{\text{policy}}} \\
& + \underbrace{\EE_{(s,s') \sim \zeta, \zeta \sim \expert} \kld(p_{\alpha,\psi}(a|s,s')|| \pi_\theta(a|s))}_{\Lcal_{\text{reg}}}
\end{split}
\label{eq:final-obj}
\end{equation}
There are no gradients flowing back to the inverse model from the $\Lcal_{\text{reg}}$ loss term. We give the complete training procedure in \refalg{alg:main}.

\subsection{Discussion on Inverse Dynamics Disagreement}\label{sec:dis_idd}
\citet{Yang2019ImitationLF} quantified the disparity between the learning objectives used within adversarial imitation learning from observations and demonstrations.  They demonstrated that this disparity is quantified by the Inverse Dynamics Disagreement (IDD),  which measures the disagreement between the inverse dynamics models of the expert and the learning agent.
\begin{equation}
\begin{split}
&\underbrace{\kld \left( \invom || \inveom \right)}_{\text{IDD}} = \\
&\underbrace{\kld \left( \saom || \saeom \right)}_{\text{ILD}} - \underbrace{\kld \left( \ssom || \sseom \right)}_{\text{ILO}}
\end{split}
\label{equ:idd_def_gap}
\end{equation}
Thus, minimizing the \ild~objective can be seen as jointly minimizing the learning objective of \ilo~and the IDD between the inverse dynamics models of the expert and the learning agent. Obviously, we have no way to access the inverse model of the expert, e.g. by learning a proxy of it, since we do not observe the expert's actions.

\begin{algorithm}[h]
  \caption{\mytitle~(\mytitleshort)}
  \begin{algorithmic}\label{alg:main}
    \REQUIRE Expert observations $\mathcal{D}_E=\{\zeta_i^E\}$ where $\zeta_i=\{ (s_0^E, s_1^E), (s_1^E, s_2^E), ...\}$, policy $\pi_\theta$, (discriminator $D_\phi$), inverse dynamics model $p_{\alpha,\psi}(a|s,s')$, a replay buffer $\Rcal$ and maximum number of iterations $M$
    \STATE Initialize replay buffer $\Rcal$
    \STATE Initialize network parameters $\theta, (\phi), \alpha, \psi$
    \FOR{$1$ to $M$}
    \STATE Collect agent rollouts $\Dcal_A=\{\xi^{\pi_{\theta}}_i\}$, $\xi_i \thicksim \pi_{\theta}$, $\xi_i=\{ (s_0, a_0, r_0, s_1), (s_1, a_1, r_1, s_2), ...\}$, with $r_i$ from \reftab{tab:reward}
    \STATE Add $\Dcal_A$ to the replay buffer $\Rcal$
    \REPEAT
    \STATE Sample uniformly a minibatch $\Bcal$ of state-action-state triplets from $\Rcal$, $(s,a,s')\sim \Rcal$
    \STATE Update the inverse dynamics model $p_{\alpha,\psi}(a|s,s')$
    \UNTIL{Inverse dynamics model $p_{\alpha,\psi}$ converges}
    %\STATE Update policy $\pi_\theta$ using $\Lcal_{\rm final}$% $\Lcal_{ppo} + \lambda \Lcal_{BCO}$
    \STATE Update policy $\pi_\theta$ using $\Lcal$ from~\refeq{eq:final-obj}
    \STATE (Update the discriminator $D_\phi$)
    \ENDFOR
  \end{algorithmic}
\end{algorithm}

Our regularizer, as defined in \refeq{eq:reg}, minimizes the KL divergence between the inverse dynamics model of the environment, $\rho(a|s,s')$, and the learned policy, $\policy(a|s)$, within the support of the state-state transitions observed from the expert.
This KL divergence is an upper bound of the KL divergence between the IDMs of the environment and that of the agent, $\invom$, (refer to Appendix \refsec{proof:rel_pol_invpol} for the proof), i.e:
\begin{equation}
\begin{split}
&\kld \left( \rho(a|s,s')  || \invom \right) \leq \\
&\kld \left( \rho(a|s,s') || \policy(a|s) \right) + \text{Const}
\end{split}
\end{equation}
Thus, at optimality of \refeq{eq:reg}, the inverse models of the environment and that of the learned agent align and we have $\rho(a|s,s') = \invom$.
This correspondence leads the IDD gap to be equal to the KL divergence between the environment's and the expert's inverse models:
\begin{equation}
\begin{split}
&\kld \left( \invom || \inveom \right) = \\
&\kld \left( \rho(a|s,s') || \inveom \right)
\end{split}
\end{equation}

We cannot manipulate any components on the right-hand side; the environment's inverse model is governed by the simulator's physics and the expert's inverse model is dictated by the expert's behavior, over which we have no control.
The environment's inverse model provides all conceivable actions that could lead to a given state-state transition under the simulator's physics. Conversely, the expert's inverse model is more selective, i.e. has smaller support, as it assigns non-zero probability only to actions the expert would have chosen for a specific state-state transition.

While ideally, we would want to minimize the IDD between the learned agent's and the expert's inverse models, the lack of access to the expert's actions leaves us with no expert-specific choices. We chose to guide the agent towards actions that are physically plausible by deploying our regularizer. This guidance offers valuable supervision, compensating for the low-quality reward signal provided by the discriminator during early stages of adversarial training. Importantly, the training of the environment's inverse model is a supervised process, which is considerably less complex and faster to converge than adversarial training.

\section{Experiments}\label{sec:exp}

\begin{figure*}[!t]
    \centering
    \begin{subfigure}{\linewidth}
        \includegraphics[width=\linewidth]{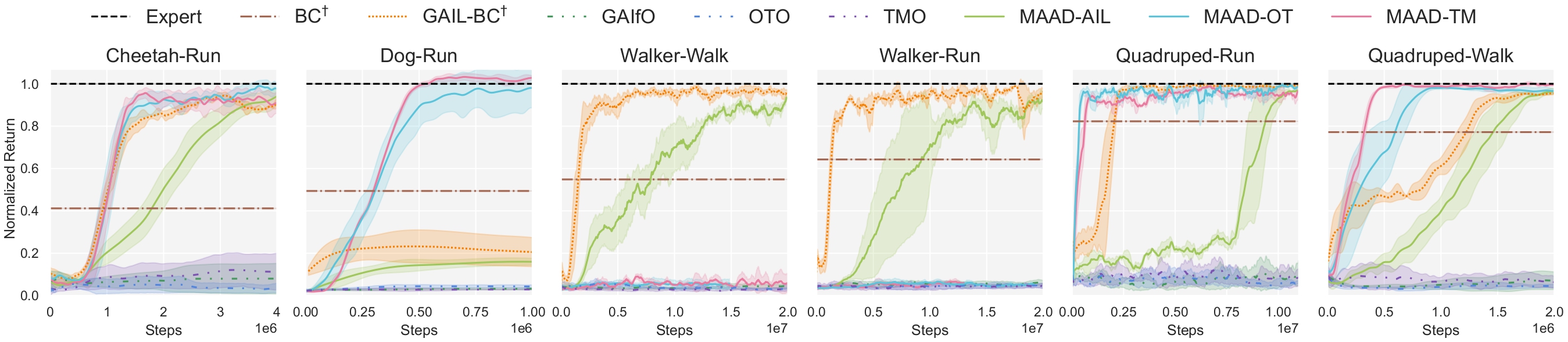}
        % \vskip -0.05in
        \caption{DMC Suite}
        \label{fig:dmc}
    \end{subfigure}
    \begin{subfigure}{\linewidth}
        \includegraphics[width=\linewidth]{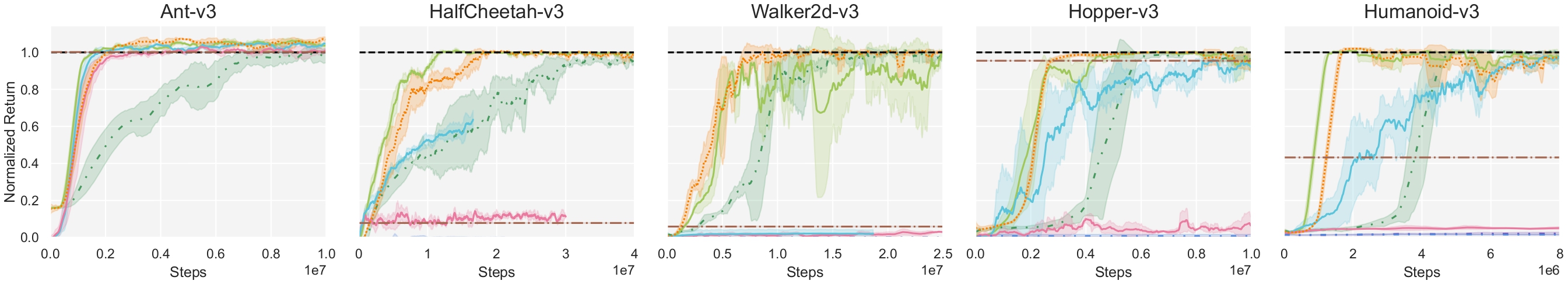}
        % \vskip -0.05in
        \caption{OpenAI Gym}
        \label{fig:gym}
    \end{subfigure}
    % \vskip -0.1in
    \caption{Performance comparison between our proposed version of \mytitleshort~and the baselines (some of the baselines, highlighted here with $\dag$, have access to expert actions, \refsec{sec:baselines} for more information). We average the results over three random seeds and show the mean and the range of one standard deviation.}
  \label{fig:results}
% \vskip -0.2in
\end{figure*}

To evaluate MAAD, we have conducted experiments on complex control tasks from the MuJoCo suite of environments \citep{todorov2012mujoco}.
We have used 6 locomotion tasks from DeepMind Control Suite and 5 from OpenAI Gym. The two suites differ on various aspects, such as their initial state distributions and termination criteria, posing their specific challenges to the learning algorithms, more details in \refsec{sec:mujoco}.
We collected expert trajectories from a policy trained using PPO \cite{Schulman2017ProximalPO} on each MuJoCo task.
Then we used the collected trajectories to train several imitation learning baseline models and compare them against different flavors of our model.
In the baseline models, we included not only \ilo~methods but also some \ild~ones which have access to expert actions, these later ones should be seen as the best possible achievable performance if actions were observed. More details about environments, hyperparameters and training data and implementation can be found in \refsec{sec:specifications} in appendix, and our code is openly available:\href{https://github.com/jacr13/MAAD}{https://github.com/jacr13/MAAD}.

\subsection{MAAD variants and Baselines} \label{sec:baselines}
We evaluate three instantiations of \mytitleshort, which differ on how surrogate rewards are obtained. We compare each one of them against its non-regularizer variant, where the KL regularizer is switched off. 
The first instantiation, \mytitleshort-AIL, obtains surrogate rewards by adversarial imitation learning; we compare it against two variants of GAIL \cite{Ho2016GenerativeAI}: GAIfO \citep{Torabi2018GenerativeAI}, which operates on observations and is essentially the non-regularized variant of \mytitleshort-AIL, and GAIL-BC, which has access to expert actions.
%, a hybrid that merges GAIL with BC to benefit from both methods. Importantly, GAIL-BC requires access to expert actions.
The second instantiation, \mytitleshort-TM, obtains rewards relying on a trajectory matching approach; we employ the Euclidean distance metric for trajectory comparisons. 
%This method is contrasted with a trajectory matching from observations (TMO) approach, similar to our \mytitleshort-TM but excluding the regularization component.
%We compare \mytitleshort-TM against its non-regularized variant which 
We denote its non-regularized variant by TMO, for trajectory matching from observations. 
The third instantiation, \mytitleshort-OT, is based on rewards sourced from optimal transport techniques. Specifically, it uses the Sinkhorn algorithm with a cosine distance-based cost matrix, akin to the frameworks of SIL and ROT. 
%We compare it against its non-regularized variant which we denote by OTO, i.e. optimal transport from observations.
We denote its non-regularized variant by OTO, i.e. optimal transport from observations. 
%However, it excludes pretraining and regularization, as these elements necessitate access to expert actions. We refer to its non-regularised variant as OTO, i.e. optimal transport from observations.

%Our comparative analysis encompasses a comprehensive set of baseline algorithms, which serve as a standard for assessing the performance of our novel approaches. 
In addition to the non-regularized MAAD baselines, we also include a behavioral cloning baseline (BC) \cite{Pomerleau91, Bain1995AFF}, which employs a supervised learning approach to learn the policy and requires action access. Although it is a fast training method since it does not require interactions with the environment, its asymptotic performance is suboptimal unless a substantial amount of expert data is available. We have also experimented with BCO \citep{TorabiWS18}, the \ilo~variant of BC, which was consistently inferior to BC. Consequently, we decided not to include it in our reported results.

\subsection{Results} \label{sec:results}
%We conduct a comparative analysis of the various instantiations of our method, \mytitleshort-*, against the baselines, focusing on sample efficiency, 
We focus our comparative analysis of the different methods on sample efficiency, i.e. the number of environmental interactions required to achieve expert performance. We provide results for the DMC suite in \reffig{fig:dmc} and for the OpenAI Gym in \reffig{fig:gym}. The \mytitleshort~variants consistently surpassed their non-regularized counterparts, underscoring the importance of guidance and their aptitude in learning in the absence of expert actions. In the appendix~\refsec{sec:detail_results}, we present detailed results, including a table summarizing the final performance and plots comparing the performance of the different models using either environment interactions or time.

The performance of the different \mytitleshort~variants varied based on evaluation suite, with some excelling in certain environments. In the DMC suite (\reffig{fig:dmc}), the TM and OT variants of MAAD had very similar convergence rates, outperforming their non-regularized counterparts (TMO and OTO) as well as the MAAD-AIL variant, the latter in most but not all environments.
Notably, the non-regularized variants, OTO and TMO, failed to converge within the allocated interaction budget, highlighting the critical role of guiding the policy even when using plausible actions as a substitute for actual expert actions.
Among the AIL-based approaches, only GAIL-BC and MAAD-AIL achieved convergence to expert performance levels, GAIfO consistently underperformed. GAIL-BC systematically outperformed MAAD-AIL; however, it was comparable or slightly slower than MAAD-OT and MAAD-TM. 
It is noteworthy that in walker-based environments, specifically in run and walk tasks, only GAIL-BC and MAAD-AIL managed to converge to expert performance within the allotted budget. All TM- and OT-based methods struggled to learn in these environments. In \refsec{sec:actions}, we compare learned and expert actions, highlighting the challenges in walker-based environments.

In the OpenAI Gym suite (\reffig{fig:gym}), the scenario differs somewhat. While the non-regularized variants, OTO and TMO, follow a similar pattern to those in the DMC suite, i.e. they do not converge within the allocated number of interactions, now their regularized versions also face challenges in converging. For instance, MAAD-TM only converges in the Ant-v3 environment. MAAD-OT does achieve expert performance, but its rate of convergence is slower than that of the adversarial versions. In this suite, walker-based environments also appear to be particularly challenging for the OT and TM variants, as none of them managed to converge.
Regarding the AIL variants, all three were capable of matching expert performance, with GAIL-BC and MAAD-AIL consistently outperforming GAIfO, demonstrating up to four times greater sample efficiency. A somewhat unexpected outcome was that \mytitleshort-AIL also slightly outperformed GAIL-BC,
%. As \mytitleshort-AIL can be seen as the \ilo~variant of GAIL-BC, one might anticipate slightly inferior performance, 
given GAIL-BC's access to expert actions. 
%However, our model utilizes the inverse model of the environment, guiding the policy towards selecting plausible actions by matching the policy with the inverse model using KL divergence.
One possible explanation for that is that the inverse model on which MAAD relies to regularize the policy provides a broader range of, plausible, actions to choose from, offering more flexibility, compared to what happens in GAIL-BC where the regularizer guides the policy to exactly match the expert actions. Such a flexibility can be important in particular at the beginning of training, as matching expert actions precisely might be more challenging.
% In contrast, GAIL-BC employs a BC regularizer to explicitly constrain the agent to mimic expert actions using a mean square error loss.
% Matching two distributions instead of matching a single target value naturally accounts for the variance inherent in the action distribution and could explain the slightly better performance.
% An alternative explanation here could be that matching for plausible actions, instead of expert actions, allows for more leeway and is an easier task, especially in the initial phases of the training.
Overall, MAAD-AIL is the method that achieves expert performance in both suites and all environments, except for the dog-run task in the DMC suite.
 
\mytitleshort~outperformed all the \ilo~baselines across various control tasks by exhibiting systematically faster convergence to expert performance, or in other terms, greater sample efficiency. Notably, for a good number of environments, a number of \ilo-baselines did not even start converging. 
Its ability to learn effectively without expert actions makes it a powerful tool for tackling imitation learning from observations problems in real-world scenarios where expert states are hard and expensive to label accurately with action.

\section{Conclusion}

We presented \mytitle~(\mytitleshort), a novel framework for imitation learning from observations. 

The novelty of our approach lies in the integration of an inverse dynamics model in the on-policy imitation learning from observations setting. The inverse model provides access to the posterior distribution of physically plausible actions for a given state-state transition and serves as an auxiliary guide for the policy, providing the latter with meaningful action suggestions despite the absence of expert actions.

Our model combines the strengths of on-policy algorithms with behavioral cloning, effectively utilizing the benefits of both to speed-up learning. We integrate these components into a unified objective function that encourages the policy to mimic the expert's state occupancy measure while also aligning with physically plausible actions as these are provided by the learned inverse model.

We  empirically validated \mytitleshort~on a number of challenging tasks, %from the MuJoCo suite, 
demonstrating superior sample efficiency against all tested baselines. Notably, our method is able to achieve expert performance in settings in which some of the baselines do not even start training, and even matches the performance of baselines with access to expert actions.

\section*{Acknowledgements}
This work was partially supported by the Swiss National Science Foundation grant number CSSII5\_177179 ``Modeling pathological gait resulting from motor impairment''.

\section*{Impact Statement}

This paper presents work whose goal is to advance the field of Reinforcement Learning, specifically in the domain of imitation learning from observations, by introducing a regularizer that encourages the agent to reproduce physically-plausible expert actions (estimated via a learned inverse dynamics model). Our research was conducted using simulators and expert data collected from a trained agents, with no real-world data involved. While our work may have various societal implications, we do not believe any particular consequences need to be highlighted here.

% \section*{Acknowledgements}
%   This work was supported by the Swiss National Science Foundation grant number CSSII5\_177179 ``Modeling pathological gait resulting from motor impairment''.

\bibliography{bibliography}
\bibliographystyle{icml2024}

%%%%%%%%%%%%%%%%%%%%%%%%%%%%%%%%%%%%%%%%%%%%%%%%%%%%%%%%%%%%%%%%%%%%%%%%%%%%%%%
%%%%%%%%%%%%%%%%%%%%%%%%%%%%%%%%%%%%%%%%%%%%%%%%%%%%%%%%%%%%%%%%%%%%%%%%%%%%%%%
% APPENDIX
%%%%%%%%%%%%%%%%%%%%%%%%%%%%%%%%%%%%%%%%%%%%%%%%%%%%%%%%%%%%%%%%%%%%%%%%%%%%%%%
%%%%%%%%%%%%%%%%%%%%%%%%%%%%%%%%%%%%%%%%%%%%%%%%%%%%%%%%%%%%%%%%%%%%%%%%%%%%%%%
\newpage
\appendix
\onecolumn
\section{Proofs}

\subsection{Connecting Policy and Environment's IDM through KL Divergence}\label{proof:rel_pol_invpol}

In this section, we present the proof that our regularizer, as defined in \refeq{eq:reg}, targets the minimization of the KL divergence between the inverse dynamics model of the environment, denoted as $\rho(a|s,s')$, and the learned policy, $\policy$, over the support of the observed state-state transitions of the expert. More specifically, we show that this KL divergence is an upper bound on the divergence between the environment's inverse dynamics model (IDM) and the agent's IDM, represented as $\invom$. Therefore, we establish that:
\begin{equation}
  \kld \left( \rho(a|s,s')  || \invom \right)  \leq \kld \left( \rho(a|s,s') || \policy(a|s) \right) + \text{Const}
\end{equation}
\begin{proof}
  \begin{align*}
     & \kld \left( \rho(a|s,s') || \invom \right)                                                                                                                                                                                                               \\
     & = \int_{\Scal\times\Acal\times\Scal} \rho(s,a,s') \log \frac{\rho(a|s,s')}{\invom}dsdads'                                                                                                                                                                \\
     & = \int_{\Scal\times\Acal\times\Scal} \rho(s,a,s') \log \frac{\rho(a|s,s')}{\underbrace{\frac{\mathcal{T}(s'|s, a)\policy}{\int_\mathcal{A} \mathcal{T}(s'|s, \bar{a})\pi_\theta(\bar{a}|s)d\bar{a}}}_{\text{By def. \refeq{eq:inversemodel}}}}dsdads'                                                                \\
     & = \int_{\Scal\times\Acal\times\Scal} \rho(s,a,s') \log \frac{\rho(a|s,s')\int_\mathcal{A} \mathcal{T}(s'|s, \bar{a})\pi_\theta(\bar{a}|s)d\bar{a}}{\policy\mathcal{T}(s'|s, a)}dsdads'                                                                         \\
     & = \int_{\Scal\times\Acal\times\Scal} \rho(s,a,s') \log \frac{\rho(a|s,s')}{\policy}dsdads' + \int_{\Scal\times\Acal\times\Scal} \rho(s,a,s') \log \frac{\int_\mathcal{A} \mathcal{T}(s'|s, \bar{a})\pi_\theta(\bar{a}|s)d\bar{a}}{\mathcal{T}(s'|s, a)}dsdads' \\
     & \leq \kld \left( \rho(a|s,s') || \policy \right) + \sup_{\rho_\pi}\left( \int_{\Scal\times\Acal\times\Scal} \rho(s,a,s') \log \frac{\int_\mathcal{A} \mathcal{T}(s'|s, \bar{a})\pi_\theta(\bar{a}|s)d\bar{a}}{\mathcal{T}(s'|s, a)}dsdads' \right)             \\
     & = \kld \left( \rho(a|s,s') || \policy \right) + \text{Const}
  \end{align*}
\end{proof}
The second term in the inequality, $\sup(.)$, is not subject to optimization with respect to the parameterized policy. Therefore, it can be regarded as a constant and we need only minimize the first term of the derived upper bound, i.e.~$\kld \left( \rho(a|s,s') || \policy \right)$.

\clearpage
\section{Specifications}\label{sec:specifications}
In this section, we give an account of the MuJoCo environments used in our experiments, the relevant aspects of our model's implementation and the chosen hyperparameters.

\subsection{Environments}\label{sec:mujoco}
We explored two suites, each with its own specificities. While OpenAI Gym employs a narrow initial state distribution, unnormalized rewards, and a termination signal when the agent falls, the DeepMind Control (dmc) suite utilizes a more challenging starting distribution, normalized rewards (i.e., each reward is in the range [0,1]), and a time limit termination of 1000 steps.
Each of these suites presents distinct challenges to the learning process. OpenAI Gym's simpler initial conditions and early termination policy tend to simplify training by limiting the exploration space. However, it employs non-normalized rewards, which may be more difficult to interpret. On the other hand, the DMC suite starts with a much larger initial state space and only terminates after the agent has performed 1000 timesteps in the environment. This significantly increases the complexity of exploration due to the greater number of possibilities. It utilizes normalized rewards, which tend to be easier for algorithms to learn from.

\reftab{table:env} provides a description of the state and action spaces of MuJoCo environments, along with the number and length of expert trajectories used to train our models. 
We use OpenAI Gym and the DMC suite as standard APIs to communicate with MuJoCo.
For the OpenAI Gym environments, the versions associated with the name correspond to the version of the environment used in the Gym library.

It is important to note that AIL-based methods utilize only a subset of these trajectories. Following the approach in the original GAIL implementation \cite{Ho2016GenerativeAI}, we employ a subsampling rate of 20. This means that we retain only 50 state-action pairs (or state-state pairs, depending on whether the model requires actions or the next state) per trajectory. As a result, the number of expert samples used is effectively reduced from from $16000$ to $800$. In contrast, all other models utilize the full set of expert trajectories.

\begin{table}[!h]
  \begin{center}
    \begin{tabular}{ l c c c }
      % \toprule
      Environment & $\Scal$     & $\Acal$  & Expert Trajectories \\
                  &             &          & $\#$ $\times$ Length \\
      \midrule
      OpenAI Gym & \multicolumn{3}{c}{}\\
          \qquad Hopper-v3   & $\RR^{11}$  & $\RR^{3}$ & 16 $\times$ 1000     \\
          \qquad HalfCheetah-v3 & $\RR^{17}$  & $\RR^{6}$ & 16 $\times$ 1000  \\
          \qquad Walker2d-v3   & $\RR^{17}$  & $\RR^{6}$ & 16 $\times$ 1000   \\
          \qquad Ant-v3      & $\RR^{111}$ & $\RR^{8}$ & 16 $\times$ 1000    \\
          \qquad Humanoid-v3 & $\RR^{376}$ & $\RR^{17}$ & 16 $\times$ 1000   \\
      \midrule
      DMC Suite & \multicolumn{3}{c}{}\\
        \qquad Cheetah-Run & $\RR^{17}$ & $\RR^{6}$ & 16 $\times$ 1000   \\
        \qquad Walker-Walk & $\RR^{24}$ & $\RR^{6}$ & 16 $\times$ 1000   \\
        \qquad Walker-Run & $\RR^{24}$ & $\RR^{6}$ & 16 $\times$ 1000   \\
        \qquad Quadruped-Run & $\RR^{78}$ & $\RR^{12}$ & 16 $\times$ 1000   \\
        \qquad Quadruped-Walk & $\RR^{78}$ & $\RR^{12}$ & 16 $\times$ 1000   \\
        \qquad Dog-Run & $\RR^{223}$ & $\RR^{38}$ & 16 $\times$ 1000   \\
      % \bottomrule
    \end{tabular}
  \end{center}
  \caption{Description of MuJoCo environments and respective experts}\label{table:env}
\end{table}

\subsection{Implementation Details}\label{sec:imp_details}

We implemented all the algorithms investigated and reported in PyTorch, maintaining a similar structure and keeping the same hyperparameters as much as possible. We used PPO \cite{Schulman2017ProximalPO} as the underlying reinforcement learning algorithm.

All the online models, which require interactions with the environment, utilize 4 parallel workers for data collection and policy updates. These models share their computed gradients before optimization and receive the average gradients from all workers for policy updating. Moreover, we ran every experiment on the same set of 3 random seeds: ${0,1,2}$.

In imitation learning, access to expert trajectories is essential. We obtained these trajectories by training a policy using PPO (with the same architecture as the evaluated models) until convergence. At convergence, we generated 16 trajectories using this policy. These generated trajectories were saved and used to train all of our imitation learning models.

As noted in \refsec{sec:mujoco}, the adversarial imitation approaches use a subsample of the trajectories (50 samples per trajectory). This subsampling is achieved by randomly sampling state-action (or state-state) pairs from each expert trajectory. In contrast, all other methods, including Behavioral Cloning (BC), trajectory matching, and optimal transport-based approaches, have access to the entire expert trajectories.

For all the algorithms tested, the policy network consists of a two-layer MLP with 128 or 256 hidden units. The policy networks predicts only the mean of the action distribution as a function of the state, while the learned action variance is state-independent. The value and discriminator networks (when needed) adopt the same architecture as the policy. The inverse dynamics model also learns the variance independent of the state, which means it is set as parameters independent of the input.

\reftab{table:hyperparameters} povide further details about the parameters used for the different algorithms.

\subsection{Hyperparameters}
\reftab{table:hyperparameters} provides a comprehensive list of the hyperparameters used for each of the evaluated algorithms in \refsec{sec:exp}.

\begin{table}[!h]
    \centering
    \begin{tabular}{c  c  c}
     \hline
      & Parameter & Value \\
     \hline
     Shared &  Batch size & 64\\
     & Rollout length & 2048\\
    &  Discount $\gamma$ & 0.99\\
    &  $\pi$ architecture & \{MLP [128,128], MLP [256,256]\}\\
    &  $\pi$ Learning rate &$10^{-4}$\\
     &  $\pi$ updates & \{3,6,9\} \\
     &  GAIL $\lambda_{\text{entropy}}$ &$0$\\
    &  PPO $\epsilon$ & \{0.1, 0.2\}\\
    &  GAE $\lambda$ &0.95\\
     &  Activation & $\tanh$\\
     &  Clip norm  & 0.5 \\
     &  Gradient penalty & 10 \\
\hline     
  AIL-based  &  $\Dcal$ architecture & MLP [128,128]\\
     &  $\Dcal$ Learning rate & $10^{-4}$\\
     &  $\Dcal$ updates & 1 \\
\hline
OT-based & Reward Scale Factor & 20 \\
        & Sinkhorn \# iterations & 100 \\
        & Sinkhorn $\epsilon$ & 0.01 \\
\hline
 IDM &  $\Rcal$ size  & $10^{5}$ \\
     &  IDM architecture  & MLP [128]\tablefootnote{In our implementation, we don't train an MLP for the variance. Instead, it is set as parameters of the network, independent of the input.}\\
     & IDM Learning Rate  & $10^{-4}$\\
     & IDM $K$ & $1$\\
\hline
     BC & Epochs & 200\\
\hline
     GAIL-BC$\dag$ & $\lambda_{\text{reg}}$ & \{1, 10\} \\
\hline
     \mytitleshort-* & $\lambda_{\text{reg}}$ & \{1, 10\} \\
     \hline
    \end{tabular}
    \caption{Hyperparameters used for different algorithms, parameters in \{\} were submited to a sweep.}
    \label{table:hyperparameters}
\end{table}

\clearpage
\section{Detailed Results}\label{sec:detail_results}
In \reftab{table:gym_res} and \reftab{table:dmc_res}, we detail the performance achieved with the learned policies. For each policy, we calculate the mean and the standard deviation across 50 generated trajectories. To facilitate visualization and comparison, we have highlighted all values that fall within 10\% of the expert's performance by rendering them in bold.
\begin{table}[h]
    \centering
\begin{tabular}{l||ccccc}
Model & Hopper-v3 & HalfCheetah-v3 & Walker2d-v3 & Ant-v3 & Humanoid-v3 \\
\toprule
Expert &  3749 $\pm$ 31 & 11802 $\pm$ 172 & 7597 $\pm$ 64 & 6269 $\pm$ 132 & 7588 $\pm$ 34 \\
\midrule
GAIL-BC$\dag$ &  \textbf{3832 $\pm$ 6} & \textbf{12038 $\pm$ 790} & \textbf{7993 $\pm$ 26} & \textbf{6612 $\pm$ 625} & \textbf{7821 $\pm$ 6} \\
GAIfO &  \textbf{3842 $\pm$ 12} & \textbf{11403 $\pm$ 2136} & \textbf{7709 $\pm$ 660} & \textbf{6310 $\pm$ 1372} & \textbf{7776 $\pm$ 9} \\
TMO &  19 $\pm$ 0 & -58 $\pm$ 85 & 1 $\pm$ 1 & -185 $\pm$ 414 & 190 $\pm$ 54 \\
OTO &  41 $\pm$ 1 & 243 $\pm$ 223 & 22 $\pm$ 8 & -337 $\pm$ 607 & 120 $\pm$ 37 \\
\midrule
MAAD-AIL &  \textbf{3822 $\pm$ 36} & 10506 $\pm$ 874 & \textbf{7537 $\pm$ 435} & \textbf{6655 $\pm$ 95} & \textbf{7557 $\pm$ 346} \\
MAAD-TM &  1544 $\pm$ 897 & 2331 $\pm$ 1089 & 360 $\pm$ 257 & \textbf{6205 $\pm$ 1448} & 480 $\pm$ 88 \\
MAAD-OT &  \textbf{3713 $\pm$ 188} & 7552 $\pm$ 1154 & 211 $\pm$ 86 & \textbf{6333 $\pm$ 1142} & \textbf{7652 $\pm$ 321} \\
\bottomrule
\end{tabular}
\caption{Learned policy performance for OpenAI Gym tasks. We report the mean and the standard deviation across 50 generated trajectories (some of the baselines, highlighted here with $\dag$, have access to expert actions, \refsec{sec:baselines} for more information).}\label{table:gym_res}
\end{table}

\begin{table}[h]
    \centering
\begin{tabular}{l|cccccc}
Model & Cheetah-Run & Walker-Walk & Walker-Run & Quadruped-Run & Quadruped-Walk & Dog-Run \\
\toprule
Expert &  663 $\pm$ 121 & 932 $\pm$ 21 & 606 $\pm$ 34 & 748 $\pm$ 25 & 888 $\pm$ 63 & 382 $\pm$ 105 \\
\midrule
GAIL-BC$\dag$ &  \textbf{692 $\pm$ 59} & \textbf{906 $\pm$ 139} & \textbf{592 $\pm$ 104} & \textbf{757 $\pm$ 34} & \textbf{885 $\pm$ 67} & 99 $\pm$ 60 \\
GAIfO &  91 $\pm$ 19 & 77 $\pm$ 24 & 37 $\pm$ 18 & 112 $\pm$ 159 & 100 $\pm$ 159 & 17 $\pm$ 4 \\
TMO &  98 $\pm$ 34 & 60 $\pm$ 31 & 38 $\pm$ 14 & 88 $\pm$ 157 & 125 $\pm$ 176 & 13 $\pm$ 3 \\
OTO &  62 $\pm$ 21 & 74 $\pm$ 29 & 36 $\pm$ 14 & 95 $\pm$ 116 & 97 $\pm$ 136 & 17 $\pm$ 5 \\
\midrule
MAAD-AIL &  \textbf{704 $\pm$ 39} & 823 $\pm$ 230 & \textbf{569 $\pm$ 154} & \textbf{745 $\pm$ 37} & \textbf{875 $\pm$ 72} & 99 $\pm$ 36 \\
MAAD-TM &  \textbf{629 $\pm$ 179} & 133 $\pm$ 95 & 50 $\pm$ 20 & \textbf{755 $\pm$ 57} & \textbf{879 $\pm$ 73} & \textbf{408 $\pm$ 76} \\
MAAD-OT &  \textbf{646 $\pm$ 145} & 63 $\pm$ 42 & 46 $\pm$ 23 & \textbf{744 $\pm$ 90} & \textbf{875 $\pm$ 76} & \textbf{398 $\pm$ 96} \\
\bottomrule
\end{tabular}
\caption{Learned policy performance for DMC Suite tasks. We report the mean and the standard deviation across 50 generated trajectories (some of the baselines, highlighted here with $\dag$, have access to expert actions, \refsec{sec:baselines} for more information).}\label{table:dmc_res}
\end{table}

\subsection{Interactions-based Comparison}
\begin{figure}[H]
  \centering
    \begin{subfigure}[b]{\linewidth}
        \includegraphics[width=\linewidth]{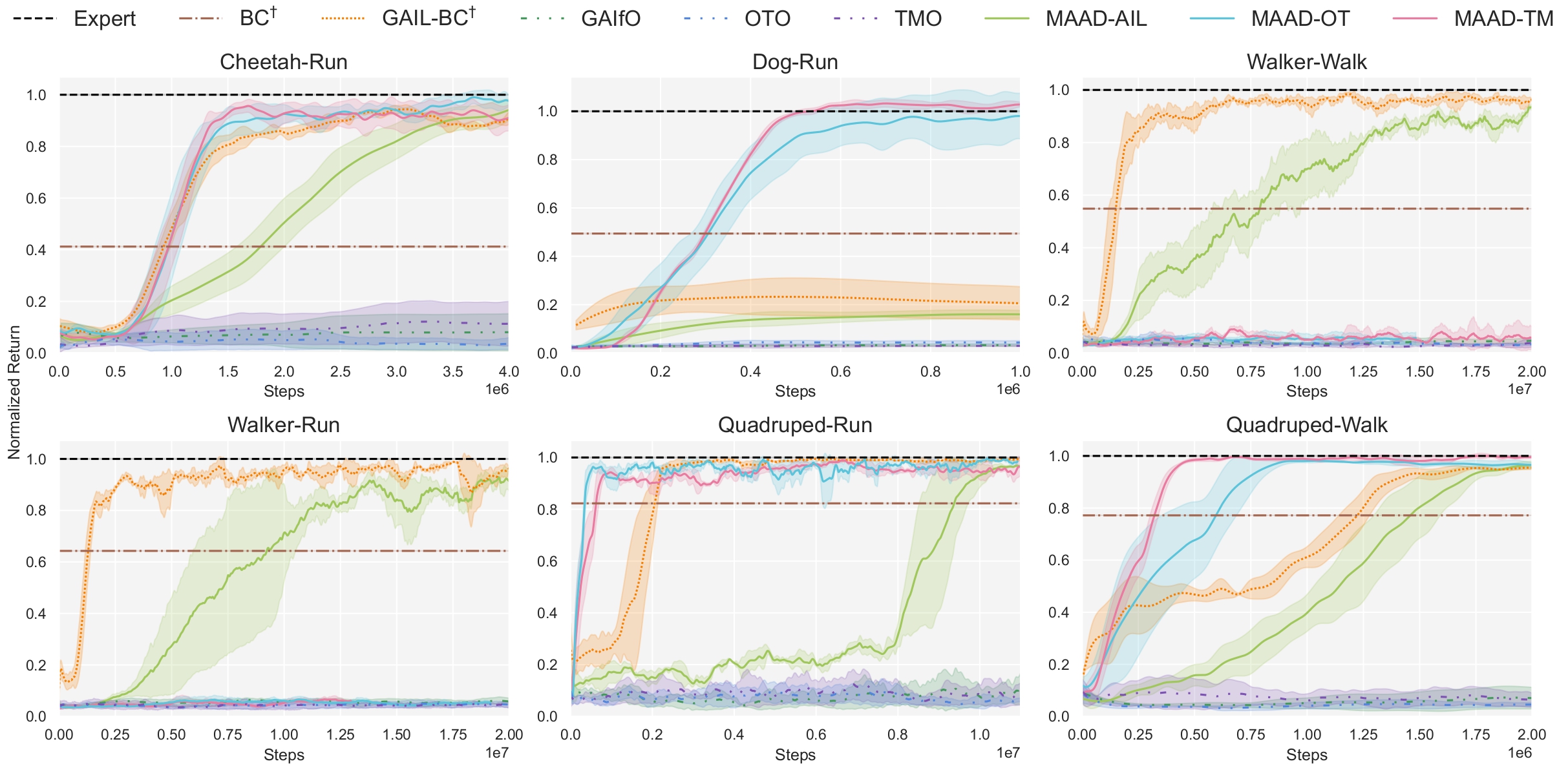}
        \caption{DMC Suite}
        \label{fig:dmc_big}
    \end{subfigure}
    \begin{subfigure}[b]{\linewidth}
        \includegraphics[width=\linewidth]{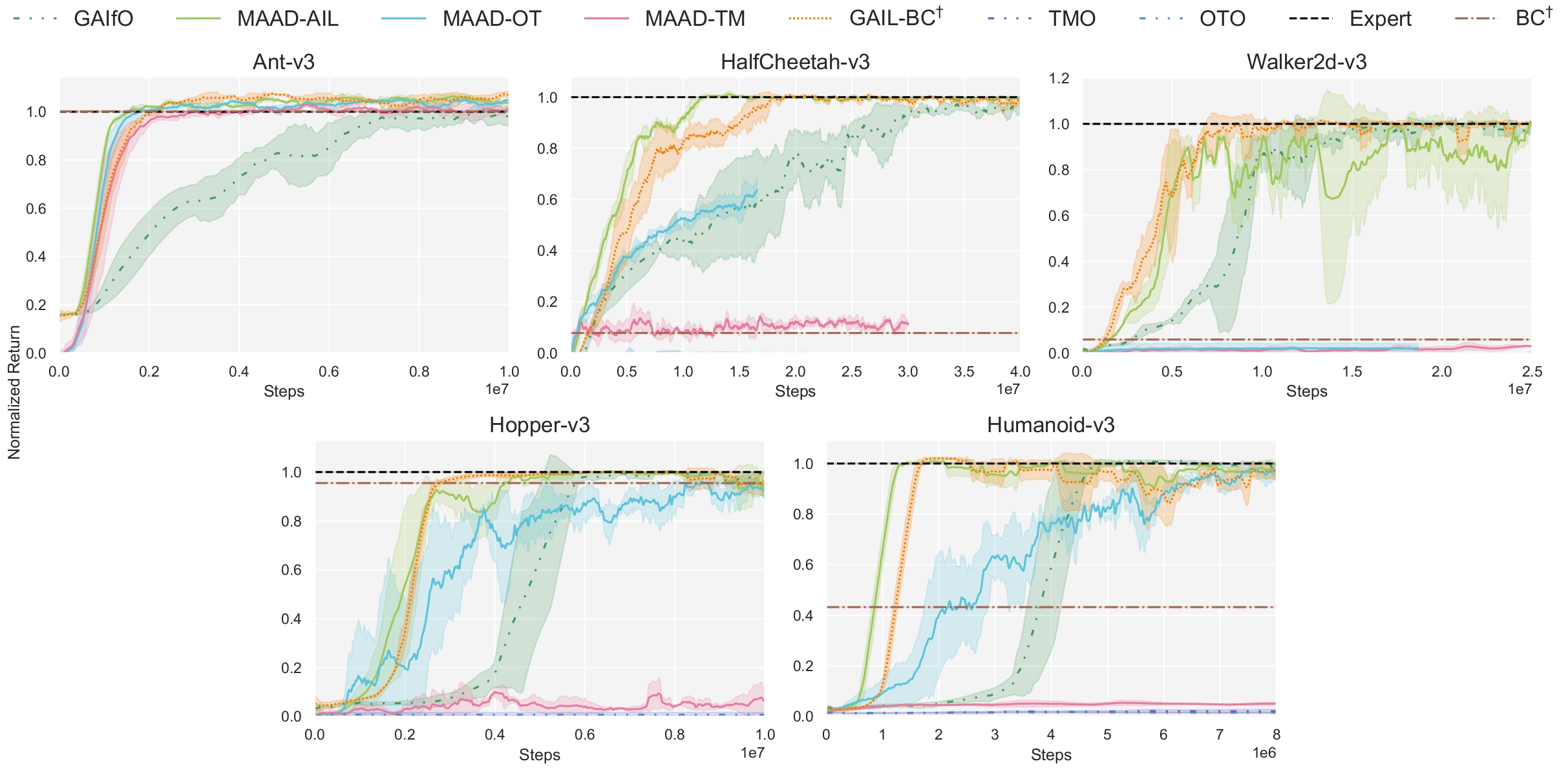}
        \caption{OpenAI Gym}
        \label{fig:gym_big}
    \end{subfigure}
    \caption{Interactions-based performance comparison of the different methods. 
    %between our proposed version of \mytitleshort~and the baselines (some of the baselines, 
    Methods marked with $\dag$, have access to expert actions, \refsec{sec:baselines} for more information). We average the results over three random seeds and show the mean and the range of one standard deviation.}
  \label{fig:results_inter_big}
\end{figure}

\subsection{Time-based Comparison}
\begin{figure}[H]
  \centering
    \begin{subfigure}[b]{\linewidth}
        \includegraphics[width=\linewidth]{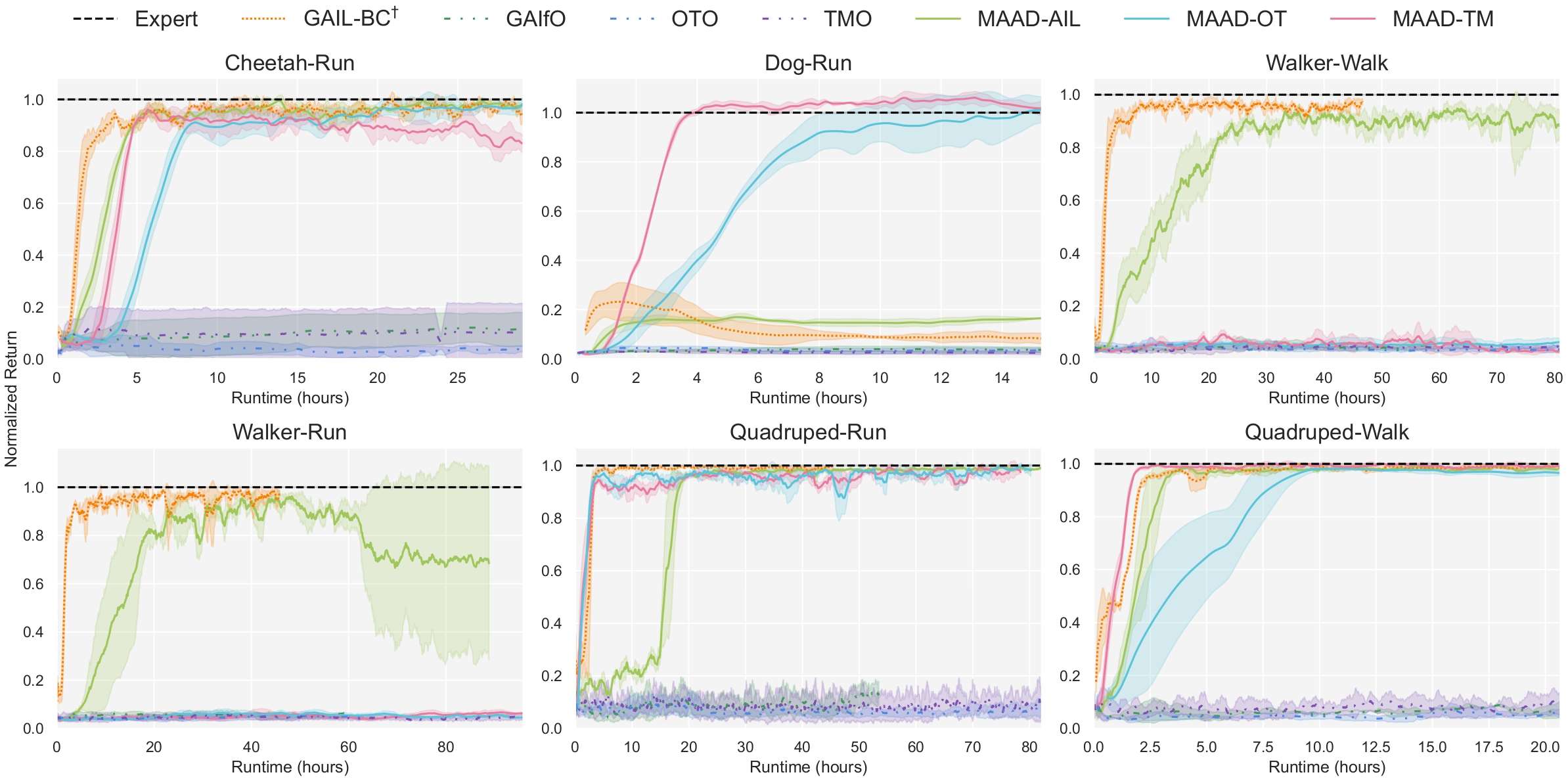}
        \caption{DMC Suite}
        \label{fig:dmc_big}
    \end{subfigure}
    \begin{subfigure}[b]{\linewidth}
        \includegraphics[width=\linewidth]{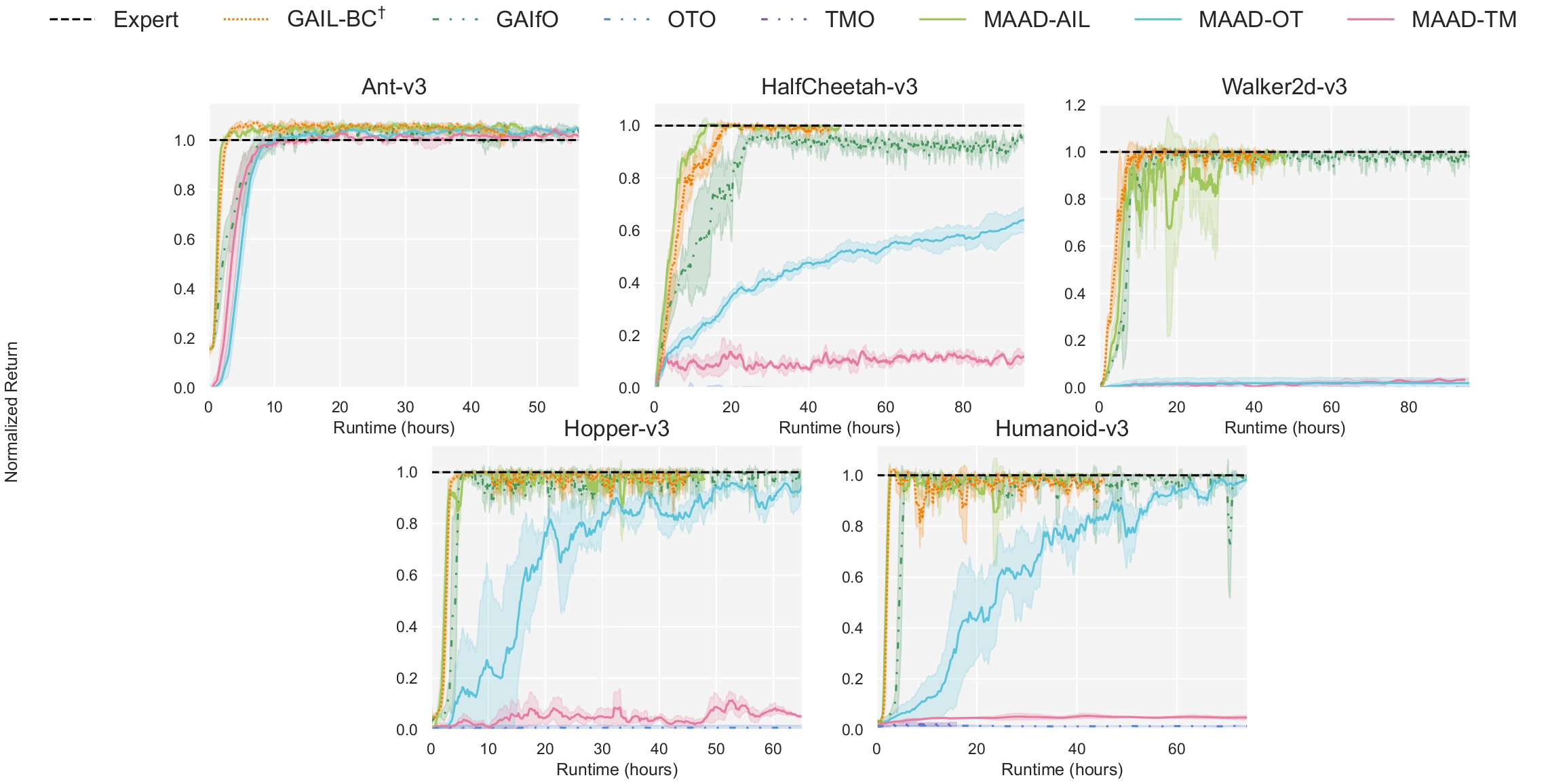}
        \caption{OpenAI Gym}
        \label{fig:gym_big}
    \end{subfigure}
    \caption{Computational time-based performance comparison of the different methods.
    %between our proposed version of \mytitleshort~and the baselines (some of the baselines, 
    Methods marked with $\dag$, have access to expert actions, \refsec{sec:baselines} for more information). We average the results over three random seeds and show the mean and the range of one standard deviation.}
  \label{fig:results_time_big}
\end{figure}

\subsection{Median Normalized Return Comparison}\label{sec:median_norm}
\begin{figure}[H]
    \centering
    \begin{subfigure}{\linewidth}
        \includegraphics[width=\linewidth]{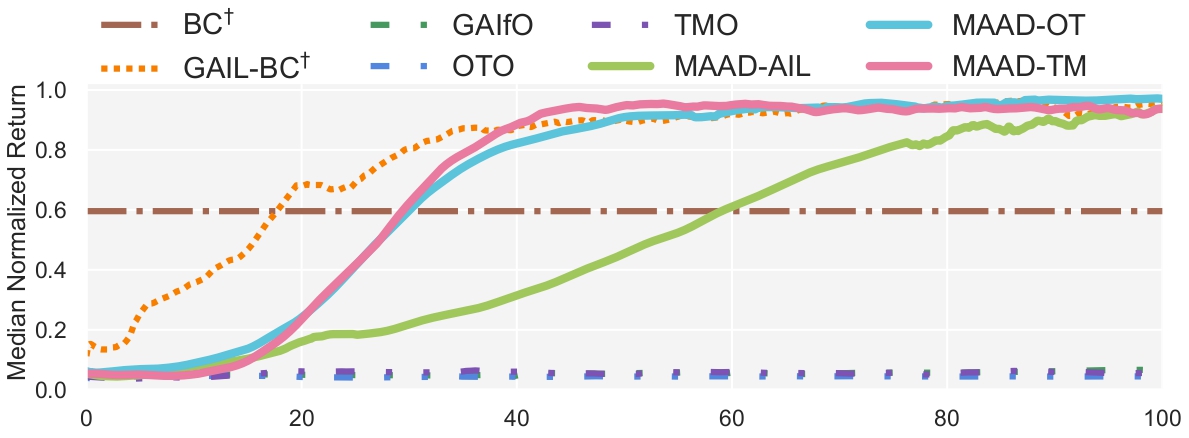}
        \caption{DMC Suite}
        \label{fig:median_dmc}
    \end{subfigure}
    \begin{subfigure}{\linewidth}
        \includegraphics[width=\linewidth]{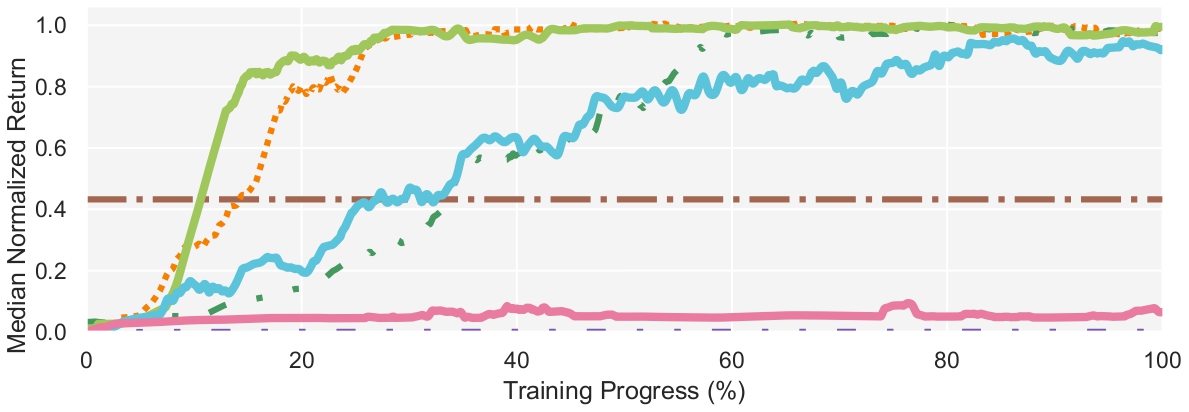}
        \caption{OpenAI Gym}
        \label{fig:median_gym}
    \end{subfigure}
    \caption{\textbf{Median Normalized Return}, over different environments, of various instantiations of our method (solid lines) versus baselines (dashed curves). 
    This plot is derived by quantalising the training curves present in \reffig{fig:results_inter_big}, using a fixed number of quantiles, here 5000, with median computation per algorithm across the different environments of each suite tested.
    Methods marked with $\dag$ have access to expert actions representing the best possible performance, all others \emph{do not}.
    }
    \label{fig:median_dmc}
    \vskip -0.2in
\end{figure}

\clearpage
\section{Comparative Analysis of Learned and Expert Actions}\label{sec:actions}

\begin{table}[H]
    \centering
\begin{tabular}{llcc}
\toprule
Env & Algo &  $\operatorname{R}^2(\pi_\theta (a|s^E), a^E)$  & $\operatorname{R}^2(\operatorname{IDM}(a|s^E_t, s^E_{t+1}), a^E)$ \\
\midrule
\multirow[t]{3}{*}{Ant-v3}
& MAAD-AIL & 0.8998 $\pm$ 6.87e-03 & 0.9492 $\pm$ 7.20e-03 \\
& MAAD-TM & 0.9350 $\pm$ 4.76e-03 & 0.9621 $\pm$ 4.64e-03 \\
& MAAD-OT & 0.9255 $\pm$ 1.05e-02 & 0.9530 $\pm$ 9.89e-03 \\
\cline{1-4}
\multirow[t]{3}{*}{HalfCheetah-v3}
& MAAD-AIL & 0.9952 $\pm$ 2.82e-04 & 0.9991 $\pm$ 1.78e-04 \\
& MAAD-TM & 0.8719 $\pm$ 5.23e-02 & 0.8762 $\pm$ 5.29e-02 \\
& MAAD-OT & 0.9916 $\pm$ 1.62e-03 & 0.9956 $\pm$ 1.65e-03 \\
\cline{1-4}
\multirow[t]{3}{*}{Walker2d-v3}
& MAAD-AIL & 0.8297 $\pm$ 2.62e-02 & 0.8575 $\pm$ 2.62e-02 \\
& MAAD-TM & \color{red}0.0192 $\pm$ 2.34e-01 & \color{red}0.0238 $\pm$ 2.35e-01 \\
& MAAD-OT & \color{red}-0.5949 $\pm$ 3.30e-01 & \color{red}-0.5816 $\pm$ 3.31e-01 \\
\cline{1-4}
\multirow[t]{3}{*}{Hopper-v3}
& MAAD-AIL & 0.7943 $\pm$ 1.13e-01 & 0.8072 $\pm$ 1.06e-01 \\
& MAAD-TM & 0.5033 $\pm$ 1.66e-01 & 0.5196 $\pm$ 1.64e-01 \\
& MAAD-OT & 0.9223 $\pm$ 2.53e-02 & 0.9269 $\pm$ 2.59e-02 \\
\cline{1-4}
\multirow[t]{3}{*}{Humanoid-v3}
& MAAD-AIL & 0.9614 $\pm$ 3.01e-03 & 0.9956 $\pm$ 8.67e-05 \\
& MAAD-TM & 0.9233 $\pm$ 4.92e-03 & 0.9944 $\pm$ 1.07e-03 \\
& MAAD-OT & 0.9782 $\pm$ 2.04e-03 & 0.9958 $\pm$ 2.26e-05 \\

\midrule
\midrule

\multirow[t]{3}{*}{Cheetah-Run}
& MAAD-AIL & 0.9500 $\pm$ 4.00e-03 & 0.9920 $\pm$ 4.18e-03 \\
& MAAD-TM & 0.9809 $\pm$ 1.39e-03 & 0.9898 $\pm$ 1.44e-03 \\
& MAAD-OT & 0.9823 $\pm$ 4.26e-03 & 0.9913 $\pm$ 3.80e-03 \\
\cline{1-4}
\multirow[t]{3}{*}{Dog-Run}
& MAAD-AIL & 0.9517 $\pm$ 1.32e-03 & 0.9966 $\pm$ 3.01e-03 \\
& MAAD-TM & 0.9700 $\pm$ 2.17e-04 & 0.9985 $\pm$ 1.61e-05 \\
& MAAD-OT & 0.9730 $\pm$ 1.12e-03 & 0.9984 $\pm$ 3.86e-05 \\
\cline{1-4}
\multirow[t]{3}{*}{Walker-Walk}
& MAAD-AIL & 0.6387 $\pm$ 3.70e-03 & 0.7145 $\pm$ 1.13e-02 \\
& MAAD-TM & \color{red}-0.1460 $\pm$ 2.34e-01 & \color{red}-0.1461 $\pm$ 2.34e-01 \\
& MAAD-OT & \color{red}-0.1965 $\pm$ 2.96e-01 & \color{red}-0.1955 $\pm$ 3.12e-01 \\
\cline{1-4}
\multirow[t]{3}{*}{Walker-Run}
& MAAD-AIL & 0.7835 $\pm$ 2.52e-02 & 0.8025 $\pm$ 2.44e-02 \\
& MAAD-TM & \color{red}-0.1207 $\pm$ 1.69e-01 & \color{red}-0.1209 $\pm$ 1.69e-01 \\
& MAAD-OT & \color{red}-0.1794 $\pm$ 1.93e-01 & \color{red}-0.1796 $\pm$ 1.92e-01 \\
\cline{1-4}
\multirow[t]{3}{*}{Quadruped-Run}
& MAAD-AIL & 0.8200 $\pm$ 4.31e-04 & 0.9995 $\pm$ 8.24e-06 \\
& MAAD-TM & 0.8947 $\pm$ 2.82e-03 & 0.9978 $\pm$ 7.60e-04 \\
& MAAD-OT & 0.8901 $\pm$ 1.90e-03 & 0.9972 $\pm$ 4.89e-04 \\
\cline{1-4}
\multirow[t]{3}{*}{Quadruped-Walk}
& MAAD-AIL & 0.9261 $\pm$ 2.94e-03 & 0.9994 $\pm$ 7.14e-05 \\
& MAAD-TM & 0.9787 $\pm$ 1.55e-03 & 0.9994 $\pm$ 2.38e-06 \\
& MAAD-OT & 0.9582 $\pm$ 5.39e-04 & 0.9994 $\pm$ 1.04e-05 \\

\bottomrule
\end{tabular}
    \caption{Comparison of R-squared Scores for Policy and Inverse Dynamics Model Against Expert Actions Across Different Environments and Algorithms}
    \label{tab:my_label}
\end{table}

Even though we operate within the Imitation Learning from Observations (\ilo) framework, where trained models do not have direct access to expert actions, we possess these actions since we trained the experts. We would like to understand better how the actions that MAAD trained policies select, as well as these that the inverse model selects, relate to the expert's actions. As already discussed in Section \ref{sec:dis_idd}, the inverse model learns the distribution of plausible actions, if this is done well, the expert's action will have a distribution that has a support that is a subset of the support of the plausible actions' distribution. To understand the relations described above, we computed the R-squared score between the actions produced by our learned policies and the true expert actions, as well as between the learned IDM and the expert actions. Values below an R-squared score of 0.5 were highlighted in red. The majority of our models achieved an R-squared score close to one, i.e. most often there is a high to very high agreement between the learning policy, the inverse model, and the expert's choices. Again, we want to stress that one should not interpret that, in the general case, as the inverse model eventually learning the expert; we rather believe that for the majority of the environments that we consider here the set of plausible actions is rather constrained, after all we have seen that the unimodality assumption ($k=1$ for IDM Section \ref{sec:idm})  works best in most of the environments; thus the expert actions can only fall within this rather constrained unimodal set of plausible actions. 
Under such conditions, it is expected that the actions learned by both the policy and the inverse model would align closely with those of the expert.

A notable exception to this high agreement pattern are the walker-based environments, where we observed the poorest performance, in some cases resulting even in negative R-squared values. 
%This pattern suggests certain complexities specific to these environments. 
One possible explanation is that these environments are not strictly unimodal, leading to difficulties in action learning. This is something that we want to investigate further by focusing on more challenging environments which feature a less constrained set of plausible actions. 
%Further investigation into the environmental characteristics and model adjustments might be necessary to address these challenges and improve model performance in such settings. We reserve this investigation into walker-based environment challenges for future work.

%%%%%%%%%%%%%%%%%%%%%%%%%%%%%%%%%%%%%%%%%%%%%%%%%%%%%%%%%%%%%%%%%%%%%%%%%%%%%%%
%%%%%%%%%%%%%%%%%%%%%%%%%%%%%%%%%%%%%%%%%%%%%%%%%%%%%%%%%%%%%%%%%%%%%%%%%%%%%%%

\end{document}